# Complexity of Scrambling:
# A New Twist to the Competence - Performance Distinction [1]

Aravind K. Joshi
Department of Computer and Information Science
University of Pennsylvania
Philadelphia, PA 19104
joshi@linc.cis.upenn.edu

**ABSTRACT**

In this paper we discuss the following issue: How do we decide whether a certain property of language is a competence property or a performance property? Our claim is that the answer to this question is not given a-priori. The answer depends on the formal devices (formal grammars and machines) available to us for describing language. We discuss this issue in the context of the complexity of processing of center embedding (of relative clauses in English) and scrambling (in German, for example) from arbitrary depths of embedding.

What properties of language are considered as belonging to 'competence' and what properties to 'performance' is not something given a-priori.

Let us assume a property P which says that center embedding (of relative clauses in English, e.g.,., *the rat the cat chased ate the cheese* ) beyond level one is very difficult to process for humans, i.e., it is not just that the performance degrades beyond one level of embedding but rather that somehow whatever the processing device is, it simply 'crashes' beyond one level of embedding[2].

Should this property be part of competence in the sense that P is a property of the grammar which characterizes competence? If P can be shown to follow from the formal characterization of the grammar (or the associated formal device or automaton) then we may as well say that the grammar itself has the property P in the sense that the grammar assigns linguistically adequate structural descriptions to all sentences up to one level of embedding and beyond one level of embedding the grammar just fails to do so. i.e., it simply 'crashes'.

The accepted wisdom is that we should not require P to follow from the properties of the grammar, i.e., P is a property of performance. How do we arrive at this conclusion? The argument goes somewhat as follows.

1. If the grammar is a finite state grammar (FSG) which allows for exactly one level of embedding then trivially P is a property of the grammar. However there are two problems here. (a) A FSG does not give us linguistically meaningful structural descriptions. The rules of a FSG are of the form $A \rightarrow bC$, where $A$ and $C$ are nonterminals and $b$ is a terminal symbol. Thus for the sentence *The dog likes ice cream* we will have a structural description of the form (Fig. 1) using rules: $S \rightarrow the XP, XP \rightarrow dog\ YP, YP \rightarrow likes\ ZP$, and $ZP \rightarrow icecream$. $XP$ and $YP$ are not well motivated linguistically. Hence, FSGs are descriptively inadequate. (b) Even if one accepts FSGs as linguistically adequate, there is still a problem. The requirement that a FSG should allow only one level of embedding is quite arbitrary. Clearly, we can construct a FSG that will accommodate center embedding up to some specified level, say m. That is, as a *class* of grammars, FSGs accommodate center embedding up to any arbitrary level. Thus the class of FSGs as a *class* does not have the property P, i.e., it is not the case that for all[3] FSGs in this class the property P holds.

---

[1] Appeared in 3e Colloque International sur les grammaires d'Arbres Adjoints (TAG+3). Technical Report TALANA-RT-94-01, TALANA, Universite' Paris 7, 1994.
The work reported here is work in progress, jointly with Tilman Becker and Owen Rambow. Since the three co-workers have not yet completely agreed on the final version of the position developed here, the present author is solely responsible for the views expressed in this paper.

[2] We could assume that the processing device crashes after level 2 or even higher, as long as it is a fixed number. What is important for our discussion is that the device crashes beyond this level. What we are disallowing is a continuous degradation of performance for arbitrary levels of embedding.

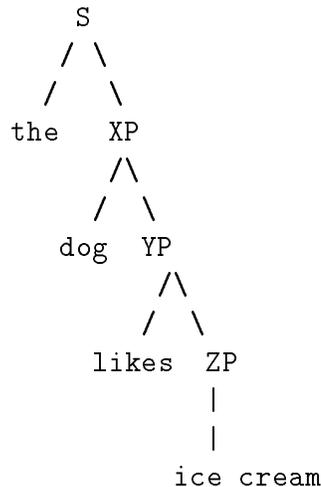

Figure 1.

2. Since FSGs are inadequate from the point of view of providing linguistically appropriate structural descriptions, we must at least consider context-free grammars (CFG). Then for the previous example (Figure 1), we can easily obtain the structural description (Figure 2.) with the associated CFG, $S \rightarrow NP\ VP, NP \rightarrow DET\ N, VP \rightarrow V\ NP, DET \rightarrow the, N \rightarrow dog, NP \rightarrow icecream$.

3. Once we leave FSGs and choose CFGs then as we know, allowing for one level of center embedding automatically allows arbitrary levels of center embedding. Putting a constraint on the level of center embedding in a CFG is quite arbitrary, something that cannot be stated in the framework of CFGs themselves. Thus, as a *class*, CFGs do not have the property P, i.e., it is not the case that for any CFG P holds.

---

[3]To be precise, *all* means *all except possibly a finite subset.*

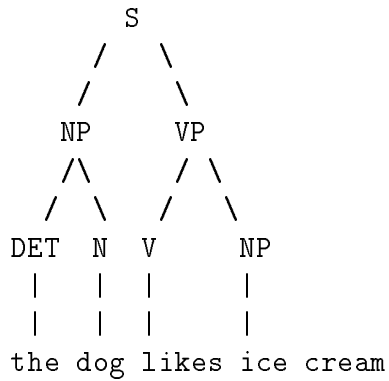

Figure 2.

4. Thus, we have reached the conclusion that it is better not to insist that P is a property of the grammar. It is better to treat P as a performance property.

Now consider an alternate scenario. Suppose there is a *class* of grammars, $G$ which have the following features: (a) they provide linguistically adequate structural descriptions, and (b) the class has the property P, i.e., each grammar in this class has the property P. Of course, to the best of our knowledge, such a class of grammars does not exist. But suppose $G$ did exist. Then it seems quite reasonable that we would adopt $G$ as it will be linguistically adequate and coincidentally capture the property P. So, why not accept $G$ and take credit for the fact that the apparent performance property P follows from the formal properties of $G$, i.e., P can be viewed as a competence property.

To summarize, given a property P whether or not P is a performance property or a competence property depends on the formal character of the grammar. If we can find a grammar $G$ (a class of grammars to be precise), which is both linguistically adequate and also captures the property P, then we adopt $G$ as our grammar and P becomes a competence property. On the other hand, if we cannot find such a grammar (a class of grammars) then we try to look for another grammar (a class of grammars, to be precise), say, $G'$ which is linguistically adequate but which does not have the property P. If we find it then we adopt $G'$ as our grammar. In this case P becomes a performance property. For the case of center embedding in English, as we have already shown, this latter situation prevails. That is, we have no choice but to choose $G'$ and treat P as a performance property. These considerations, although not expressed in this way before, as far as we know, have led to the accepted wisdom that the fact that arbitrary number of center embeddings do not really occur is to be treated as a performance constraint (property).

In this paper, we will describe a new situation that is similar on the surface to the center embedding case but actually offers us a choice that is not available in the case of center embedding in English. As far as we know, this is the first known situation where a choice is available. Thus this situation provides a sharp example for our claim that whether a property is a performance property or a competence property really depends on the formal characterization of the grammar (a class of grammars) that we have at our disposal for describing the linguistic facts. The situation we have in mind is *scrambling*, both local and long-distance, which has recently attracted considerable attention among linguists and computational linguists. We will investigate this situation in the context of tree-adjoining grammars (TAGs).

In deciding whether scrambling as a linguistic phenomenon can adequately be described by a TAG or a TAG-equivalent formalism, it is crucial to decide whether or not sentences corresponding to the strings with with 2 or more levels of embedding are indeed grammatical. Sentences involving scrambling from more than 2 levels of embedding are indeed difficult to process and native speakers do show reluctance in accepting these sentences. Now there are two directions we can follow.

- The reluctance that some native speakers show for accepting the more complex sentences (i.e., with scrambling from more than 2 levels of embedding) is due mainly to processing difficulties, rather than to the ungrammaticality of the sentences. This position is analogous to the accepted wisdom for the complexity of center embedded sentences of English. If we accept this position then we should look for further extensions of TAGs that will allow scrambling from any arbitrary level of embedding. Indeed such classes of grammars have been investigated by the authors elsewhere, especially by Rambow (1994) in his Ph.D. dissertation, in a very extensive manner.

- To follow the other direction, let us define a property Q such that a grammar (a class of grammars, to be precise) has the property Q if the grammar not only produces the scrambled strings for scrambling up to 2 levels of embedding but also assings appropriate structural descriptions satisfying the strong co-occurrence constraint and further that beyond 2 levels of embedding the grammar (the class of grammars) cannot maintain the strong co-occurrence constraint, even though it can produce the scrambled strings, i.e., the grammar *crashes* beyond 2 levels of embedding. Every grammar in this class must have this property. Suppose we can find such a class of grammars, say, $G$, then we can adopt $G$ as our grammar and claim that the property Q (which reflects the reluctance of speakers to accept sentences with scrambling from more than 2 levels of embedding) is a competence property because it is a property that follows from the formal nature of the grammar itself.

    *Note that in the case of center embedded sentences of English we do not have this kind of choice.* But now we have the choice. It can be shown that the class of tree-local MC-TAG has the property Q, i.e., every grammar in this class can handle scrambling from up to 2 levels of embedding and it will *crash* beyond 2 levels of embedding! So if adopt tree-local MC-TAG as our grammar then Q is competence property and not a performance property.

    In conclusion, we will show that the phenomenon of scrambling in the context of tree-local MC-TAG provides a unique situation allowing us to make the competence-performance distinction in a novel way.

4